\documentclass[runningheads]{llncs}

\usepackage{eccv}

\usepackage{eccvabbrv}

\usepackage{graphicx}
\usepackage{booktabs}
\usepackage{pifont}
\usepackage{multirow}
\usepackage{wrapfig}
\usepackage{paralist}

\usepackage[accsupp]{axessibility}  

\usepackage{hyperref}

\usepackage{orcidlink}

\begin{document}

\title{%
\texorpdfstring{%
Learning Human-Object Interaction\\
for 3D Human Pose Estimation\\
from LiDAR Point Clouds%
}{Learning Human-Object Interaction for 3D Human Pose Estimation from LiDAR Point Clouds}%
} 

\titlerunning{HOIL}

\author{Daniel Sungho Jung\inst{1} \and
Dohee Cho\inst{1} \and
Kyoung Mu Lee\inst{1,2}}

\authorrunning{D.S. Jung et al.}

\institute{$^1$IPAI, $^2$Dept. of ECE\&ASRI, Seoul National University \\
\email{\{dqj5182, jdh12245, kyoungmu\}@snu.ac.kr}}

\maketitle

\begin{abstract}
  Understanding humans from LiDAR point clouds is one of the most critical tasks in autonomous driving due to its close relationships with pedestrian safety, yet it remains challenging in the presence of diverse human-object interactions and cluttered backgrounds.
  Nevertheless, existing methods largely overlook the potential of leveraging human-object interactions to build robust 3D human pose estimation frameworks.
  There are two major challenges that motivate the incorporation of human-object interaction.
  First, human-object interactions introduce spatial ambiguity between human and object points, which often leads to erroneous 3D human keypoint predictions in interaction regions.
  Second, there exists severe class imbalance in the number of points between interacting and non-interacting body parts, with the interaction-frequent regions such as hand and foot being sparsely observed in LiDAR data.
  To address these challenges, we propose a Human-Object Interaction Learning (HOIL) framework for robust 3D human pose estimation from LiDAR point clouds.
  To mitigate the spatial ambiguity issue, we present human-object interaction-aware contrastive learning (HOICL) that effectively enhances feature discrimination between human and object points, particularly in interaction regions.
  To alleviate the class imbalance issue, we introduce contact-aware part-guided pooling (CPPool) that adaptively reallocates representational capacity by compressing overrepresented points while preserving informative points from interacting body parts.
  In addition, we present an optional contact-based temporal refinement that refines erroneous per-frame keypoint estimates using contact cues over time.
  As a result, our HOIL effectively leverages human-object interaction to resolve spatial ambiguity and class imbalance in interaction regions.
  Codes will be released.
\end{abstract}    
\section{Introduction}
\label{sec:intro}
Every day, pedestrians navigate sidewalks and cross roadways in close proximity to traffic, often under conditions that present significant challenges for autonomous driving systems.
They may walk, run, or interact with objects such as bicycles, electric scooters, or carried items, complicating LiDAR perception.
Ensuring pedestrian safety in such scenarios, especially when humans interact with objects, is particularly challenging.
Therefore, accurate understanding of human–object interaction is critical for reliable 3D human pose estimation from LiDAR point clouds, enabling autonomous vehicles to anticipate human behavior and operate safely and robustly in real-world environments.

\begin{figure}[t]
\begin{center}
\phantomsubcaption\label{fig:challenges:a}
\phantomsubcaption\label{fig:challenges:b}
\includegraphics[width=1.0\linewidth]{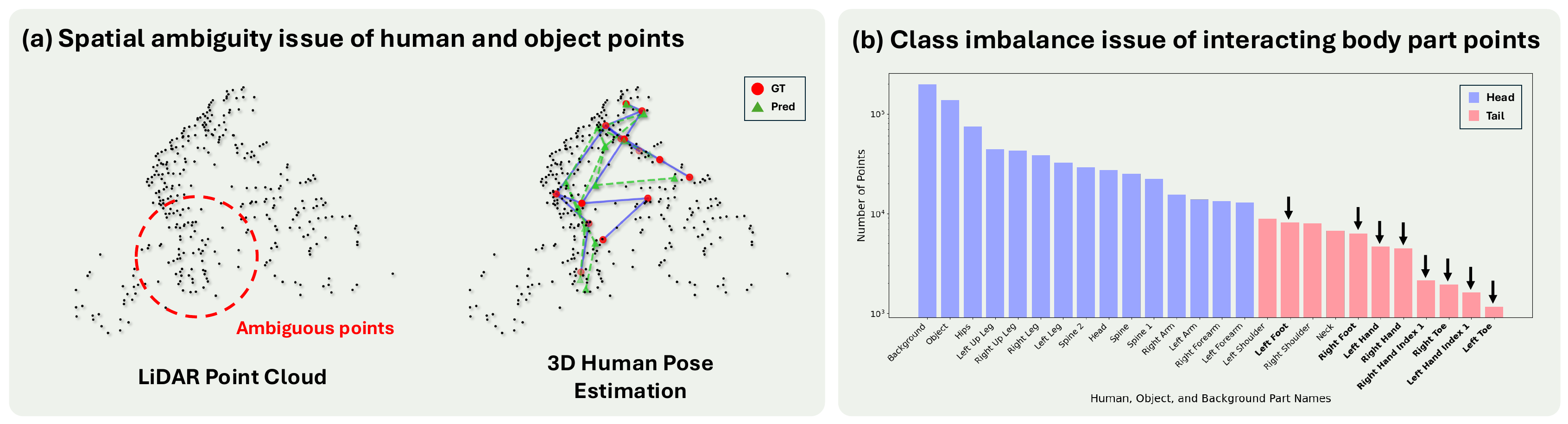}
\end{center}
\vspace{-3.5mm}
\caption{
\textbf{Main challenges for 3D human pose estimation from LiDAR points.} First, due to spatial ambiguity issue of 3D human and object points, previous SOTA method~\cite{an2025pre} fails to predict accurate pose. Second, class imbalance issue occurs in frequently interacting body parts~($\downarrow$) such as hand and foot, where the number of points are severely low compared to other body parts, object, and background.}
\label{fig:challenges}
\vspace{-0.7cm}
\end{figure}

Recent methods~\cite{weng20233d, an2025pre} predominantly follow a two-stage framework, first pre-training on SMPL~\cite{loper2015smpl}-based synthetic LiDAR data with diverse human poses to learn a body pose prior, followed by fine-tuning on a target dataset to adapt to real-world LiDAR data. 
Although this formulation leads to improvements in general scenarios, existing methods still struggle in cases involving complex interactions, particularly in predicting 3D keypoints in interacting body regions.
With the recent emergence of human–object interaction datasets~\cite{bhatnagar2022behave, jiang2023full, zhang2023neuraldome, li2023object, huang2024intercap}, we aim to expand the scope of pre-training from body pose prior learning to human–object interaction prior learning.

There are two major challenges in 3D human pose estimation from LiDAR points.
First, there exists spatial ambiguity issue between 3D human points and 3D object points.
Unlike RGB images, LiDAR points contain significantly low semantic information to distinguish between human region and interacting object region.
In Fig.~\ref{fig:challenges:a}, we can observe that, even to our eyes, it is very difficult to distinguish exactly which points are human points and which points are object points.
This shows that it is important to learn a strong prior to distinguish the difference between 3D human points and 3D object points.
Second, the class imbalance issue occurs in human and object LiDAR points.
While hand~\cite{jung2025learning} and foot~\cite{jung2025shoe} are the most frequently interacting regions (FIR) of the human body, they contain a very small number of points due to their small size relative to the entire body, resulting in significantly fewer LiDAR points than other regions.
In Fig.~\ref{fig:challenges:b}, we observe that even on a logarithmic scale, the number of LiDAR points for the hand and foot are much lower than that for other body parts.

To address these challenges, we propose HOIL, a human–object interaction learning framework that builds interaction-aware point representations for robust 3D human pose estimation from LiDAR point clouds.
During pre-training, we scale HOIL to diverse human–object interactions by leveraging the five independent human–object interaction datasets~\cite{bhatnagar2022behave, jiang2023full, zhang2023neuraldome, li2023object, huang2024intercap} in Table~\ref{tab:dataset}, enabling the model to learn an interaction prior beyond body pose prior.
To resolve spatial ambiguity between human and object points, we propose human–object interaction-aware contrastive learning~(HOICL) during pre-training. 
HOICL discriminates human and object regions via contrastive learning on the part-segmentation feature space, with emphasis on interaction and contact regions where human and object points are often mixed.
This supervision trains the model to construct discriminative feature representations for human and object points, useful when the corresponding points occupy similar spatial locations and thus are difficult to distinguish.
After pre-training, HOIL is fine-tuned on each target real-world LiDAR dataset~\cite{sun2020scalability, dai2023sloper4d} using only the 3D human pose estimation objective, leveraging the already discriminative feature space learned from large-scale human–object interaction datasets.
To tackle the severe class imbalance between interacting and non-interacting body part regions, HOIL introduces contact-aware part-guided pooling~(CPPool), which predicts pooling weights directly from the point features. 
During point encoding, CPPool reduces the contribution of dense non-contact body and object points while increasing the contribution of points from frequently interacting regions~(\textit{e.g.,} hand and foot), which contain fewer LiDAR points, ensuring that these regions remain represented after downsampling.

\begin{table}[t]
\centering
\small
\caption{\textbf{Dataset configuration.} We utilize 5 human-object interaction datasets.}
\scalebox{0.87}{\begin{tabular}{lcccccc} \toprule
Dataset & Body model & \# of samples & \# of subjects & \# of objects & \# of instructions \\
\midrule
BEHAVE~\cite{bhatnagar2022behave} & SMPL & 22K & 8 & 20 & 6 \\
CHAIRS~\cite{jiang2023full} & SMPL-X & 921K & 46 & 81 & 32 \\
HODome~\cite{zhang2023neuraldome} & SMPL-H & 632K & 10 & 23 & -- \\
OMOMO~\cite{li2023object} & SMPL-X & 5K & 17 & 15 & 103 \\
InterCap~\cite{huang2024intercap} & SMPL-X & 51K & 10 & 10 & -- \\
\bottomrule
\end{tabular}}
\vspace{-0.5cm}
\label{tab:dataset}
\end{table}

As a result, HOIL achieves state-of-the-art performance across diverse real-world LiDAR datasets for 3D human pose estimation from LiDAR point clouds, leveraging human–object interaction-aware contrastive learning and contact-aware part-guided pooling.
Our key contributions are as follows:
\raggedbottom
\begin{itemize}
\item We introduce HOIL, a novel framework for 3D human pose estimation that effectively learns human-object interaction from LiDAR point clouds.
\item To alleviate spatial ambiguity issue of human and object points, we propose human-object interaction-aware contrastive learning~(HOICL), which discriminates human and object point features. 
\item To address the class imbalance of interacting body part points, we present contact-aware part-guided pooling~(CPPool), which aggressively pools overrepresented non-contacting body parts while preserving underrepresented contacting body parts.
\item In the end, HOIL demonstrates strong performance on LiDAR point clouds involving diverse human–object interactions.
\end{itemize}
\section{Related works}
\label{sec:related_works}

\noindent\textbf{LiDAR-based 3D human pose estimation.}
With the rise of autonomous driving, accurate 3D human pose estimation from LiDAR has become an important research area. 
HPERL~\cite{furst2021hperl} proposed end-to-end pose estimation from RGB and LiDAR for accurate absolute positioning. 
Zheng~\textit{et al.}~\cite{zheng2022multi} introduced weak supervision using pseudo labels for point-wise segmentation. 
HUM3DIL~\cite{zanfir2023hum3dil} leveraged pixel-aligned multi-modal features and Transformer refinement for semi-supervised learning with 2D and 3D labels. 
FusionPose~\cite{cong2023weakly} addressed multi-person pose estimation through multi-modal fusion with self-supervised constraints. 
GC-KPL~\cite{weng20233d} proposed a fully self-supervised framework using synthetic data with SMPL~\cite{loper2015smpl} meshes. 
LPFormer~\cite{ye2024lpformer} presented an end-to-end model for joint prediction of keypoints, bounding boxes, and semantic segmentation. 
DAPT~\cite{an2025pre} decomposed LiDAR pose estimation into body prior learning and data adaptation within a Point Transformer V3~\cite{wu2024point} framework. 
Our HOIL builds on prior works~\cite{ye2024lpformer, an2025pre} by explicitly modeling human–object interaction to address two major challenges of spatial ambiguity and class imbalance.

\noindent\textbf{Human-object interaction.}
Human–object interaction~(HOI) with everyday objects such as bicycles and luggage is common in outdoor scenes perceived by LiDAR. 
Existing HOI datasets provide valuable resources for disambiguating human and object regions. 
BEHAVE~\cite{bhatnagar2022behave} captures HOIs using multi-view RGB-D data with SMPL-based fitting and contact annotations. 
CHAIRS~\cite{jiang2023full} focuses on human–chair interactions using an articulated chair model with hybrid inertial–optical motion capture, enabling analysis of seated interactions relevant to road scenarios. 
HODome~\cite{zhang2023neuraldome} records HOIs in a multi-view dome with detailed geometry, SMPL-X~\cite{pavlakos2019expressive} parameters, and object pose and shape. 
OMOMO~\cite{li2023object} provides household-object interactions captured with Luma scans and a Vicon system. 
InterCap~\cite{huang2024intercap} introduces whole-body HOIs with objects relevant to driving scenarios, such as skateboards and umbrellas.
In our HOIL framework, we use these five HOI datasets~\cite{bhatnagar2022behave, jiang2023full, zhang2023neuraldome, li2023object, huang2024intercap} to learn diverse human–object interactions from synthetic LiDAR point clouds during pre-training.

\noindent\textbf{Supervised contrastive learning.}
Cross-entropy loss alone does not explicitly enforce inter-class separability in the learned feature space~\cite{liu2016large, elsayed2018large}.
SupCon~\cite{khosla2020supervised} introduced fully supervised contrastive learning that treats all same-class samples as positives.
KCL~\cite{kang2020exploring} extended contrastive learning by sampling multiple positives from the same class using label information.
BCL~\cite{zhu2022balanced} addressed class imbalance by balancing gradient contributions to maintain well-separated class representations.
TSC~\cite{li2022targeted} promoted uniform class separation by mapping features toward predefined targets on a hypersphere.
HiMulConE~\cite{zhang2022use} extended supervised contrastive learning to hierarchical multi-label settings using ancestry-aware positive pairs.
CBL~\cite{tang2022contrastive} enhanced point cloud segmentation by improving feature discrimination across classes and scales near scene boundaries.
MulSupCon~\cite{zhang2024multi} supported multi-label classification by weighting pairs according to label overlap for fine-grained supervision.
Our HOIL leverages supervised contrastive learning in frequently interacting regions and contact regions to solve spatial ambiguity issue.
\section{Method}
\label{sec:method}

\subsection{Preliminary}
\label{subsec:preliminary}
Our architecture is based on Point Transformer V3~(PTv3)~\cite{wu2024point}, with key improvements in pooling to address class imbalance issue and learning objective to tackle spatial ambiguity issue.
Before introducing HOIL, we briefly review PTv3.

PTv3 is a Transformer~\cite{vaswani2017attention}-based hierarchical encoder-decoder point model.
Given a point cloud $\mathbf{P} \in \mathbb{R}^{N \times 3}$ containing $N$ points and $\mathbf{p}_\mathrm{n} \in \mathbb{R}^3$ denoting the $n$-th point, PTv3 first defines an ordering on the unordered point set~$\mathbf{P}$ via space-filling curve serialization~$\phi$~\cite{sagan2012space}.
Ordering the points according to serialization~$\phi$ produces serialized points~$\widetilde{\mathbf{P}}$ in which points are arranged according to locality-preserving curves~(\textit{e.g.,} Z-order~\cite{morton1966computer} or Hilbert~\cite{hilbert1935stetige}) specified by~$\phi$:
\begin{equation}
\label{eq:pre_serialization}
\widetilde{\mathbf{P}} = \mathrm{Serialize}(\mathbf{P}; \phi).
\end{equation}
The serialized points~$\widetilde{\mathbf{P}}$ are embedded by $f_{\mathrm{emb}}$ to produce point features~$\mathbf{F}_\mathrm{p} \in \mathbb{R}^{N \times C}$, where $C$ denotes feature channel dimension.
These features are then processed by a multi-stage encoder~$\mathcal{E} = \{\mathcal{E}^{(1)}, \ldots, \mathcal{E}^{(L)}\}$ for $L$ stages.
Each encoder~$\mathcal{E}^{(i)}$ in stage~$i$ applies a Grid Pooling operation~$f_{\mathrm{pool}}$, which aggregates points and features within local 3D grids using a max pooling operation to reduce spatial resolution:
\begin{equation}
\label{eq:pooling}
(\mathbf{P}^{(i)}, \mathbf{H}^{(i)}) =
f_{\mathrm{pool}}\!\left(\mathbf{P}^{(i-1)}, \mathbf{F}_\mathrm{p,enc}^{(i-1)}\right),
\end{equation}
where $(\mathbf{P}^{(i)}, \mathbf{H}^{(i)})$ denote the pooled point coordinates and features at stage~$i$.
For initial point set~$\mathbf{P}$ and features~$\mathbf{F}_\mathrm{p}$ after embedding, we denote them as 
$(\mathbf{P}^{(0)}, \mathbf{F}_\mathrm{p,enc}^{(0)})$.
Note that the Grid Pooling operation~$f_{\mathrm{pool}}$ is independent of the serialization~$\phi$ and is performed solely based on local 3D grids.
As the Grid Pooling operations~$f_{\mathrm{pool}}$ may corrupt the ordering, the pooled points~$\mathbf{P}^{(i)}$ and features~$\mathbf{H}^{(i)}$ are then reordered with the same serialization~$\phi$:
\begin{equation}
\label{eq:stage_reorder}
(\widetilde{\mathbf{P}}^{(i)}, \widetilde{\mathbf{H}}^{(i)}) =
\mathrm{Serialize}\!\left(\mathbf{P}^{(i)}, \mathbf{H}^{(i)}; \phi\right),
\end{equation}
where $(\widetilde{\mathbf{P}}^{(i)}, \widetilde{\mathbf{H}}^{(i)})$ denote the reordered coordinates and features.
Lastly, a Transformer block in stage~$i$ operates on the reordered point and feature sequence:
\begin{equation}
\label{eq:stage_transformer}
\mathbf{F}_\mathrm{p,enc}^{(i)} =
\mathcal{T}^{(i)}\!\left(\widetilde{\mathbf{P}}^{(i)}, \widetilde{\mathbf{H}}^{(i)}\right),
\end{equation}
where $\mathcal{T}^{(i)}$ denotes the Transformer block of the $i$-th encoder stage.
This procedure is repeated for all encoder stages $i = 1, \ldots, L$, producing progressively downsampled point sets and features, with the final output of encoder~$\mathcal{E}$ given by $(\mathbf{P}^{(L)}, \mathbf{F}_\mathrm{p,enc}^{(L)})$.

The PTv3 decoder~$\mathcal{D}$ mirrors the encoder~$\mathcal{E}$ and progressively restores resolution through unpooling operation~$f_{\mathrm{unpool}}$. 
Let $\mathcal{D} = \{\mathcal{D}^{(L)}, \ldots, \mathcal{D}^{(1)}\}$ denote the decoder stages corresponding to the encoder stages in reverse order. 
Unlike the encoder~$\mathcal{E}$, the decoder~$\mathcal{D}$ does not compute new spatial correspondences; instead, it reuses the pooling indices produced during encoding.
Specifically, during each encoder stage, the Grid Pooling operation~$f_{\mathrm{pool}}$ implicitly defines an assignment of fine-level points to coarse 3D grids. 
We denote by $\mathcal{M}^{(i)}$ the mapping from points at stage~$i-1$ to pooled points at stage~$i$. 
These mappings are stored and later used for feature propagation during decoding.
At decoder stage~$i$, coarse features are mapped back to the finer point set using the corresponding stored mapping:
\begin{equation}
\label{eq:unpooling}
(\widehat{\mathbf{P}}^{(i-1)}, \widehat{\mathbf{F}}_\mathrm{p}^{(i-1)}) =
f_{\mathrm{unpool}}\!\left(
\mathbf{P}^{(i)}, \mathbf{F}_\mathrm{p}^{(i)}, \mathcal{M}^{(i)}
\right),
\end{equation}
where $(\widehat{\mathbf{P}}^{(i-1)}, \widehat{\mathbf{F}}_\mathrm{p}^{(i-1)})$ denotes the points and features propagated to the finer level.
To preserve fine-grained information, the propagated features are fused with the corresponding encoder features via skip connections:
\begin{equation}
\label{eq:skip}
\mathbf{F}_{\mathrm{p},\mathrm{dec}}^{(i-1)} =
\widehat{\mathbf{F}}_\mathrm{p}^{(i-1)} \oplus \mathbf{F}_{\mathrm{p},\mathrm{enc}}^{(i-1)},
\end{equation}
where $\mathbf{F}_{\mathrm{p},\mathrm{enc}}^{(i-1)}$ denotes the features from encoder stage~$i-1$ and $\oplus$ represents feature fusion.
Since the decoder reuses the stored mappings $\mathcal{M}^{(i)}$ and does not require additional serialization or neighborhood construction, the decoding process is computationally lightweight and primarily serves to recover resolution.
After the final stage, the decoder restores the point coordinates 
and features $(\mathbf{P}^{(0)}, \mathbf{F}_{\mathrm{p},\mathrm{dec}}^{(0)})$ at the original resolution.

\subsection{Model architecture}
Given a LiDAR point cloud $\mathbf{P} \in \mathbb{R}^{N \times 3}$, we first extract point-wise features using a PTv3 backbone~\cite{wu2024point} described in Section~\ref{subsec:preliminary}.
PTv3 processes the points through a hierarchical encoder-decoder and produces point features at the original resolution, yielding final decoder features $\mathbf{F}_{\mathrm{p},\mathrm{dec}}^{(0)} \in \mathbb{R}^{N \times C}$, where $C=256$ denotes the feature channel dimension.
Our only architectural modification to PTv3 is to replace the max pooling operation in Grid Pooling $f_{\mathrm{pool}}$ (Eq.~\ref{eq:pooling}) with the proposed contact-aware part-guided pooling (CPPool) described in Section~\ref{subsec:cppool}, while all other PTv3 components remain unchanged.
Following DAPT~\cite{an2025pre}, we introduce learnable keypoint queries to represent $N_k$ human keypoints.
Let $\mathbf{Q} \in \mathbb{R}^{N_k \times C}$ denote the keypoint queries, where each query corresponds to one keypoint.
To inject point-level information into these queries, we employ a cross-attention Transformer that takes the keypoint queries $\mathbf{Q}$ as queries and the point features $\mathbf{F}_{\mathrm{p},\mathrm{dec}}^{(0)}$ as keys and values.
The attention loads spatial coordinates of the points, enabling each keypoint query to aggregate relevant spatial features from the point set and produce updated queries~$\mathbf{Q}$.
Finally, we predict outputs via four prediction heads to obtain point-level segmentation and contact along with 3D keypoint coordinates and keypoint-level contact.
From the point features $\mathbf{F}_{\mathrm{p},\mathrm{dec}}^{(0)}$, a point-level segmentation head predicts human-object part segmentation~$\mathbf{S}_\mathrm{p} \in \mathbb{R}^{N \times K}$ and a point-level contact head predicts point-level contact~$\mathbf{C}_\mathrm{p} \in \mathbb{R}^{N \times 1}$.
From the updated keypoint queries~$\mathbf{Q}$, a keypoint-level coordinate head predicts 3D keypoint coordinates~$\mathbf{K} \in \mathbb{R}^{N_k \times 3}$ and a keypoint-level contact head predicts keypoint-level contact~$\mathbf{C}_\mathrm{K} \in \mathbb{R}^{N_k \times 1}$.
Each head is implemented as a lightweight MLP.

\begin{figure}[t]
\begin{center}
\includegraphics[width=1.0\linewidth]{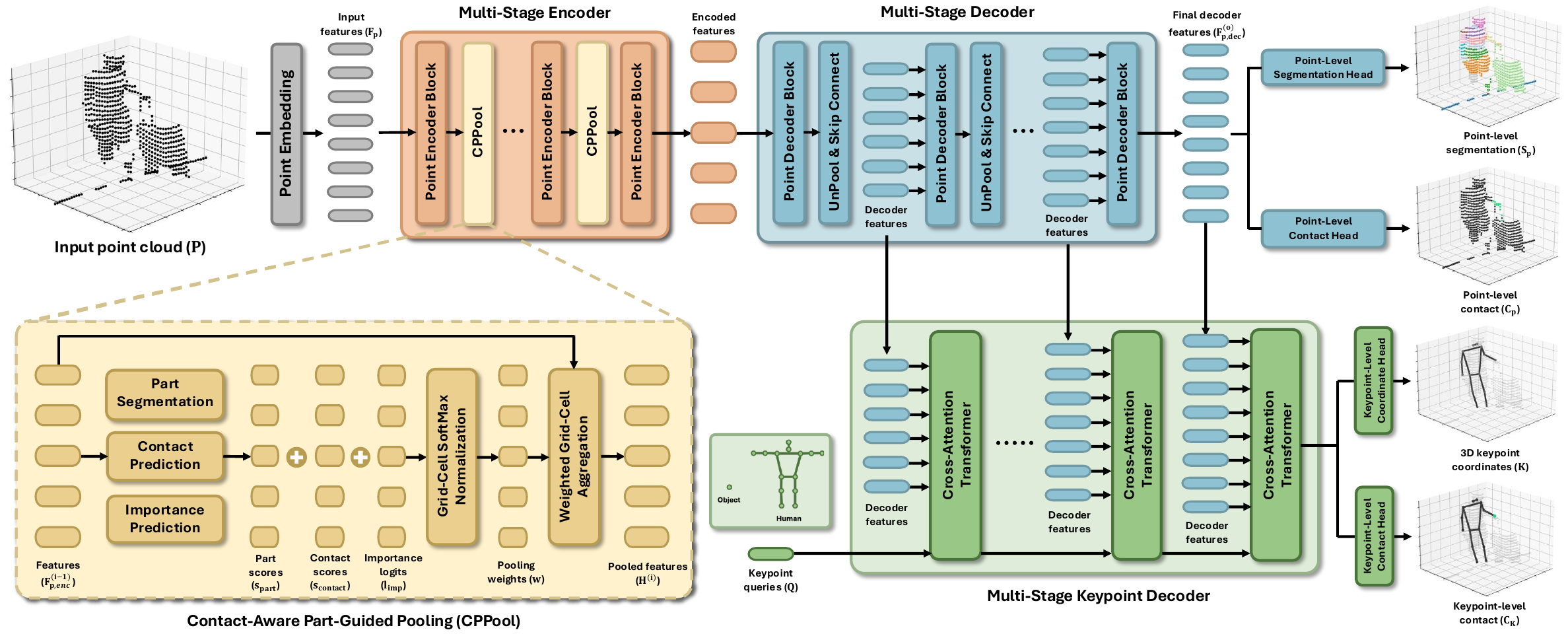}
\end{center}
\vspace{-4.5mm}
\caption{
\textbf{Overall pipeline of HOIL.} Given an input point cloud~$\mathbf{P}$, we first embed it into input features~$\mathbf{F}_\mathrm{p}$ and encode them using a multi-stage encoder with CPPool. The encoded features are then progressively decoded through a multi-stage decoder to produce the final decoder features~$\mathbf{F}_{\mathrm{p},\mathrm{dec}}^{(0)}$. At each decoding stage, keypoint queries~$\mathbf{Q}$ are iteratively updated via multi-stage keypoint decoder with decoder features. Lastly, HOIL predicts point-level segmentation~$\mathbf{S}_\mathrm{p}$ and contact~$\mathbf{C}_\mathrm{p}$ from~$\mathbf{F}_{\mathrm{p},\mathrm{dec}}^{(0)}$, and 3D keypoint coordinates~$\mathbf{K}$ and keypoint-level contact~$\mathbf{C}_\mathrm{K}$ from keypoint queries~$\mathbf{Q}$.}
\label{fig:hotl_overall_pipeline}
\vspace{-0.3cm}
\end{figure}

\subsection{Contact-aware part-guided pooling}
\label{subsec:cppool}
The grid pooling operations~(\textit{e.g.,} max pooling) from PTv3 uniformly sample points within local 3D grid cells, potentially discarding points that suffer from class imbalance issue in Fig.~\ref{fig:challenges:b}.
To address this issue, we propose contact-aware part-guided pooling~(CPPool), illustrated in Fig.~\ref{fig:hotl_overall_pipeline}, which predicts pooling weights that preserve information from interacting regions.

At encoder stage~$i$, we replace the max pooling operation from $f_{\mathrm{pool}}$ in Eq.~\ref{eq:pooling} with our CPPool applied within each local 3D grid cell.
Let $\mathbf{F}_\mathrm{p,enc}^{(i-1)} \in \mathbb{R}^{N^{(i-1)} \times C}$ denote the dense point features before pooling, corresponding to dense points $\mathbf{P}^{(i-1)}$.
CPPool computes three point-wise signals over $\mathbf{P}^{(i-1)}$: part scores $s_{\mathrm{part}}$, contact scores $s_{\mathrm{contact}}$, and importance logits $\ell_{\mathrm{imp}}$.

To predict the part scores~$s_{\mathrm{part}}$, we estimate part probabilities~$\boldsymbol{\pi}_{\mathrm{part}}$ with an auxiliary part segmentation head $f_{\mathrm{part}}$, implemented as a lightweight MLP applied to point features, followed by softmax normalization:
\begin{equation}
\boldsymbol{\pi}_{\mathrm{part}} =
\mathrm{softmax}\!\left(f_{\mathrm{part}}(\mathbf{F}_\mathrm{p,enc}^{(i-1)})\right).
\end{equation}
The part scores~$s_{\mathrm{part}}$ are computed as the dot product between the part probabilities~$\boldsymbol{\pi}_{\mathrm{part}}$ and a fixed part weight vector $\mathbf{w}_{\mathrm{part}} \in \mathbb{R}^{K}$, which assigns larger weights to frequently interacting parts such as hand and foot, formally
$s_{\mathrm{part}} =
\boldsymbol{\pi}_{\mathrm{part}} \cdot \mathbf{w}_{\mathrm{part}}$,
where $K$ is the number of part classes.

The contact score~$s_{\mathrm{contact}}$ are directly predicted by an auxiliary contact head $f_{\mathrm{contact}}$ with a lightweight MLP, followed by a sigmoid function:
\begin{equation}
s_{\mathrm{contact}} =
\mathrm{sigmoid}\!\left(f_{\mathrm{contact}}(\mathbf{F}_\mathrm{p,enc}^{(i-1)})\right).
\end{equation}

To incorporate contextual information, CPPool also predicts an importance logit~$\ell_{\mathrm{imp}}$ using point features~$\mathbf{F}_\mathrm{p,enc}^{(i-1)}$, a global feature $\mathbf{F}_\mathrm{g,enc}^{(i-1)}$, and a keypoint feature $\mathbf{F}_\mathrm{k,enc}^{(i-1)}$.
The global feature $\mathbf{F}_\mathrm{g,enc}^{(i-1)}$ is computed by average pooling point features~$\mathbf{F}_\mathrm{p,enc}^{(i-1)}$, while the keypoint feature~$\mathbf{F}_\mathrm{k,enc}^{(i-1)}$ is obtained by the learnable keypoint queries.
Please refer to supplementary material for details on keypoint queries.
These scores are fused by a lightweight MLP $f_{\mathrm{imp}}$ to produce an importance logit for each point:
\begin{equation}
\ell_{\mathrm{imp}} =
f_{\mathrm{imp}}\!\left(\mathbf{F}_\mathrm{p,enc}^{(i-1)}, \mathbf{F}_\mathrm{g,enc}^{(i-1)}, \mathbf{F}_\mathrm{k,enc}^{(i-1)}\right).
\end{equation}
The three terms are combined into a final pooling logit:
\begin{equation}
\boldsymbol{\ell} =
\frac{\ell_{\mathrm{imp}}}{T}
+ \lambda_{\mathrm{part}} \log(s_{\mathrm{part}})
+ \lambda_{\mathrm{contact}} \log(s_{\mathrm{contact}}),
\end{equation}
where $T$ is a temperature and $\lambda_{\mathrm{part}}, \lambda_{\mathrm{contact}}$ control contributions of each prior.
Let $\text{cell}(j)$ denote the set of point indices inside grid cell $j$ and ($\ell_n$, $\ell_t$) denote the ($n$-th, $t$-th) element of $\boldsymbol{\ell}$.
We apply a softmax over the pooling logits~$\ell$ of points in $\text{cell}(j)$ to obtain pooling weights $w_n^{(j)}$:
\begin{equation}
w_{n}^{(j)} =
\frac{\exp(\ell_n)}
{\sum_{t \in \text{cell}(j)} \exp(\ell_t)},
\end{equation}
where $n$ denotes the target point inside cell $j$, and $t$ runs over all points in that cell.
The weights are thus positive and sum to one within each cell.
Using these weights, CPPool computes the pooled feature~$\mathbf{H}^{(i)}j$ for each grid cell~$j$ by aggregating projected point features in that cell:
\begin{equation}
\mathbf{H}^{(i)}_j =
\sum_{n \in \text{cell}(j)}
w_{n}^{(j)} \, f_{\mathrm{proj\_pool}}\!\left(\mathbf{F}_{\mathrm{p,enc},\mathrm{n}}^{(i-1)}\right),
\end{equation}
where $f_{\mathrm{proj\_pool}}$ is a lightweight MLP applied to point features before pooling.
Stacking $\mathbf{H}^{(i)}_j$ over all grid cells forms the pooled feature tensor~$\mathbf{H}^{(i)}$.
The pooled points and features are then reordered according to serialization to form the sequence that can be consumed by the subsequent Transformer blocks:
\begin{equation}
(\widetilde{\mathbf{P}}^{(i)}, \widetilde{\mathbf{H}}^{(i)}) =
\mathrm{Serialize}\!\left(\mathbf{P}^{(i)}, \mathbf{H}^{(i)}; \phi\right),
\end{equation}
where $(\widetilde{\mathbf{P}}^{(i)}, \widetilde{\mathbf{H}}^{(i)})$ denote the reordered coordinates and features.
The Transformer block~$\mathcal{T}^{(i)}$ at stage $i$ operates on the reordered sequence to produce updated point features:
$\mathbf{F}_\mathrm{p,enc}^{(i)} = \mathcal{T}^{(i)}\!\left(\widetilde{\mathbf{P}}^{(i)}, \widetilde{\mathbf{H}}^{(i)}\right)$.
In the end, CPPool preserves sparse yet interaction-critical body part information during downsampling and solves class imbalance issue of interacting body part points.

\subsection{Human-object interaction-aware contrastive learning}
\label{subsec:hoicl}
In the regions of human-object interaction, spatial ambiguity between human and object points frequently occurs as it is difficult to distinguish human and object points as in Fig.~\ref{fig:challenges:a}.
To enhance feature discrimination in these regions, we introduce human-object interaction-aware contrastive learning (HOICL).

\begin{figure}[t]
\begin{center}
\includegraphics[width=1.0\linewidth]{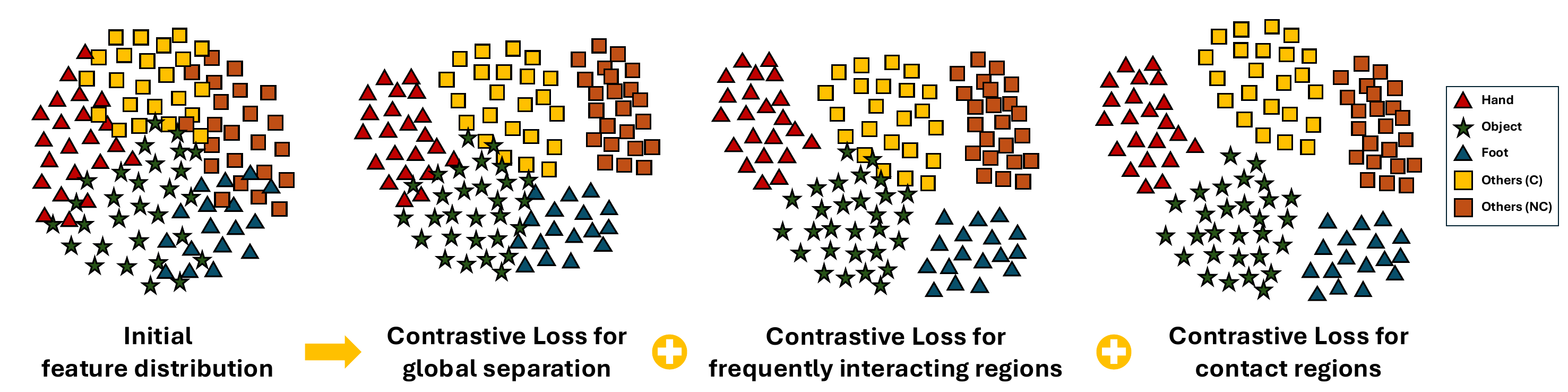}
\end{center}
\vspace{-4.5mm}
\caption{
\textbf{Human-object interaction-aware contrastive learning.} C and NC refer to contact and non-contact, respectively. Others represent other human body parts except hand and foot. For the simplicity of illustration, we do not divide hand and foot for C and NC whereas our HOICL does operate on those cases as well.}
\label{fig:ho_cl}
\vspace{-0.3cm}
\end{figure}

HOICL operates on the final decoder features at the original resolution,
$\mathbf{F}_{\mathrm{p},\mathrm{dec}}^{(0)} \in \mathbb{R}^{N \times C}$.
Let $\mathbf{f}_\mathrm{n} \in \mathbb{R}^{C}$ denote the feature of the $n$-th point in $\mathbf{F}_{\mathrm{p},\mathrm{dec}}^{(0)}$.
Each point feature~$\mathbf{f}_\mathrm{n}$ is projected into a normalized embedding space:
\begin{equation}
\mathbf{z}_\mathrm{n} =
\frac{f_{\mathrm{proj\_cl}}(\mathbf{f}_\mathrm{n})}{\| f_{\mathrm{proj\_cl}}(\mathbf{f}_\mathrm{n}) \|_2}
\in \mathbb{R}^{D},
\end{equation}
where $f_{\mathrm{proj\_cl}}$ is a lightweight MLP.
Note that as this requires point-wise ground-truth labels for part and contact, we only conduct HOICL during pre-training where synthetic LiDAR points are made from SMPL human and object meshes.

Our HOICL loss~$\mathcal{L}_{\mathrm{HOICL}}$ function consists of three components that operate at different levels: global separation, FIR to object alignment, and human to object contact alignment.
Here, FIR refers to the hand and foot.
The overall loss is defined as:
\begin{equation}
\mathcal{L}_{\mathrm{HOICL}}
= \mathcal{L}_{\mathrm{global}} + \lambda_{\mathrm{fir}}\,\mathcal{L}_{\mathrm{fir}} + \lambda_{\mathrm{hoc}}\,\mathcal{L}_{\mathrm{hoc}} .
\end{equation}
The global term enforces separability of all part features using a mix of hierarchical contrastive objective~$\mathcal{L}_{\mathrm{HMLC}}$~\cite{zhang2022use} and targeted contrastive objective~$\mathcal{L}_{\mathrm{TSC}}$~\cite{li2022targeted}:
\begin{equation}
\mathcal{L}_{\mathrm{global}}
=
\lambda_{\mathrm{hmlc}}\,\mathcal{L}_{\mathrm{HMLC}}
+
\lambda_{\mathrm{tsc}}\,\mathcal{L}_{\mathrm{TSC}} .
\end{equation}
This is to incorporate both the hierarchical structure of human-object parts while still enforcing contrastive objective for all body-object parts.
Note that parts are defined by 24 SMPL body parts and 1 part for the whole object.

To emphasize interaction regions, we introduce FIR to object contrastive learning.
Let $\mathcal{Y}_{\mathrm{fir}}$ denote a point set of frequently interacting body regions~(\textit{e.g.,} hand and foot) and $\mathcal{Y}_{\mathrm{obj}}$ denote a point set of object regions.
We apply a supervised contrastive loss~(SupCon)~\cite{khosla2020supervised} to embeddings belonging to either set.
In contrastive learning, for each anchor embedding $\mathbf{z}_n$, a positive pair $(n,m)$ is a pair encouraged to have high similarity, while a negative pair $(n,l)$ is a pair encouraged to have low similarity.
Accordingly, we define a binary mask matrix $\mathbf{M}^{\mathrm{fir}} \in \{0,1\}^{N' \times N'}$ over selected points, where $N' = |\mathcal{Y}_{\mathrm{fir}} \cup \mathcal{Y}_{\mathrm{obj}}|$ and $M^{\mathrm{fir}}_{\mathrm{nm}} = 1$ indicates that $(n,m)$ is a positive pair~(\textit{i.e.,} $\mathbf{z}_\mathrm{m}$ is treated as a positive for anchor $\mathbf{z}_\mathrm{n}$), whereas $M^{\mathrm{fir}}_\mathrm{nm} = 0$ indicates a negative pair.
In FIR, we set $M^{\mathrm{fir}}_\mathrm{nm}=1$ if points $n$ and $m$ share the same category label~(both FIR or both object), and $0$ otherwise.
Let $\tau_{\mathrm{fir}}$ denote the temperature parameter.
The loss is defined as:
\begin{equation}
\mathcal{L}_{\mathrm{fir}}
=
\mathrm{SupCon}\!\left(
\{\mathbf{z}_\mathrm{n}\}_{n \in \mathcal{Y}_{\mathrm{fir}} \cup \mathcal{Y}_{\mathrm{obj}}},
\mathbf{M}^{\mathrm{fir}};
\tau_{\mathrm{fir}}
\right).
\end{equation}

Finally, we consider human-contact points and object-contact points.
Let $\mathcal{Y}_{\mathrm{hc}}$ and $\mathcal{Y}_{\mathrm{oc}}$ denote human contact regions and object contact regions, respectively.
A SupCon loss is applied to embeddings belonging to either set.
As above, for each anchor embedding $\mathbf{z}_\mathrm{n}$, $\mathbf{M}^{\mathrm{hoc}}$ specifies which embeddings are treated as positives~(pulled closer) versus negatives~(pushed apart).
We define $\mathbf{M}^{\mathrm{hoc}} \in \{0,1\}^{N'' \times N''}$ over the selected points, where $N'' = |\mathcal{Y}_{\mathrm{hc}} \cup \mathcal{Y}_{\mathrm{oc}}|$ and
$\mathbf{M}^{\mathrm{hoc}}_\mathrm{nm} = 1$ indicates a positive pair for anchor $n$, while $\mathbf{M}^{\mathrm{hoc}}_\mathrm{nm} = 0$ indicates a negative pair.
In our contact-alignment setting, we set $\mathbf{M}^{\mathrm{hoc}}_\mathrm{nm} = 1$ if points $n$ and $m$ belong to the same category~(both human-contact or both object-contact), and $0$ otherwise.
Let $\tau_{\mathrm{hoc}}$ denote the corresponding temperature.
The loss is defined as:
\begin{equation}
\mathcal{L}_{\mathrm{hoc}} =
\mathrm{SupCon}\!\left(
\{\mathbf{z}_n\}_{n \in \mathcal{Y}_{\mathrm{hc}} \cup \mathcal{Y}_{\mathrm{oc}}},
\mathbf{M}^{\mathrm{hoc}};
\tau_{\mathrm{hoc}}
\right).
\end{equation}
In the end, HOICL tackles spatial ambiguity issue by enforcing discriminative representations between human and object points in interaction regions.

\subsection{Contact-based temporal refinement}
\label{subsec:ctrefine}
Given the estimated 3D keypoints~$\mathbf{K} \in \mathbb{R}^{N_k \times 3}$ and keypoint-level contact predictions $\mathbf{C}_\mathrm{K} \in \mathbb{R}^{N_k \times 1}$ from HOIL, we optionally further refine the keypoint estimates using a contact-based temporal refinement module.
Specifically, we first employ a self-attention Transformer solely for the keypoint coordinates~$\mathbf{K}$ to conduct temporal modeling.
Then, we employ a cross-attention Transformer in which $\mathbf{K}$ serves as query tokens, while the keys and values are constructed from the concatenation of keypoints and contact.
We add a residual connection from $\mathbf{K}$ to the output of cross-attention, and thus we produce refined keypoints $\mathbf{K}_{\mathrm{refine}}$, whose effectiveness is evaluated in Table~\ref{tab:temporal}.

\subsection{Final outputs and loss functions}

\noindent\textbf{Pre-training.}
Given the decoder features at the original resolution
$\mathbf{F}_\mathrm{p,dec}^{(0)} \in \mathbb{R}^{N \times C}$,
we use four heads to predict human-object part segmentation, point-level contact, 3D keypoints, and keypoint-level contact as final outputs.
The pre-training objective is:
\begin{equation}
\mathcal{L}_{\mathrm{pretrain}}
=
\mathcal{L}_{\mathrm{seg}}
+
\mathcal{L}_{\mathrm{contact}}
+
\mathcal{L}_{\mathrm{coord}}
+
\mathcal{L}_{\mathrm{K\_contact}}
+
\mathcal{L}_{\mathrm{HOICL}}
+
\mathcal{L}_{\mathrm{CPPool}},
\end{equation}
where segmentation loss~$\mathcal{L}_{\mathrm{seg}}$ and contact loss~$\mathcal{L}_{\mathrm{contact}}$ are CE loss, coordinate loss~$\mathcal{L}_{\mathrm{coord}}$ is MSE loss, $\mathcal{L}_{\mathrm{K\_contact}}$ is a CE loss for keypoint contact, and $\mathcal{L}_{\mathrm{HOICL}}$ is the loss function from Section~\ref{subsec:hoicl}.
We additionally apply losses to train the predictors used inside CPPool, and denote their sum as $\mathcal{L}_{\mathrm{CPPool}}$.
Specifically, $\mathcal{L}_{\mathrm{CPPool}}$ includes a segmentation loss for $f_{\mathrm{part}}$ and a contact loss for $f_{\mathrm{contact}}$, and a loss for training the importance prediction used for pooling weights.
Please refer to supplementary material for details on $\mathcal{L}_{\mathrm{CPPool}}$.

\noindent\textbf{Fine-tuning.}
During fine-tuning on real-world LiDAR datasets~\cite{sun2020scalability,dai2023sloper4d},
we follow DAPT~\cite{an2025pre} and predict 3D keypoints as final output via axis-wise 1D heatmaps.
The fine-tuning objective is:
\begin{equation}
\mathcal{L}_{\mathrm{finetune}}
=
\mathcal{L}_{\mathrm{heatmap}}
+
\mathcal{L}_{\mathrm{limb}},
\end{equation}
where heatmap loss~$\mathcal{L}_{\mathrm{heatmap}}$ supervises the predicted 1D heatmaps on the $x,y,z$ axes via a KL divergence loss, same as in DAPT,
and we add limb loss~$\mathcal{L}_{\mathrm{limb}}$~\cite{duan2022revisiting} that regularizes via joint connection to encourage kinematically consistent bones.
Please refer to supplementary material for details on $\mathcal{L}_{\mathrm{limb}}$.
\section{Implementation details}
\label{sec:implementation_details}

PyTorch~\cite{paszkepytorch} is used for implementation.
Our architecture is based on the PTv3 framework~\cite{wu2024point} and its basic configuration.
We use the AdamW optimizer~\cite{loshchilov2018decoupled} with learning rate of $3 \times 10^{-4}$ during pre-training and $5 \times 10^{-4}$ during fine-tuning, both with mini-batch size of 64.
For stable convergence, we apply a cosine annealing learning rate schedule.
Point clouds are voxelized with a grid size of 0.01.
We generate synthetic LiDAR point clouds for pre-training by ray casting onto SMPL and 3D object meshes, with additional ground and wall planes.
We train HOIL for 50 epochs, each for pre-training and fine-tuning, on a single NVIDIA A6000 GPU.
\section{Experiments}
\label{sec:experiments}

\subsection{Datasets}
We take 5 datasets with diverse human-object interactions including BEHAVE~\cite{bhatnagar2022behave}, CHAIRS~\cite{jiang2023full}, HODome~\cite{zhang2023neuraldome}, OMOMO~\cite{li2023object}, InterCap~\cite{huang2024intercap} for pre-training HOIL on human-object interactions.
To balance the number of samples across datasets, we apply sampling ratios of 1, 40, 30, 1, and 1, respectively.
Although these datasets always contain both human and object in each sample, real-world scenarios may include human alone; therefore, we randomly remove object with a ratio of 0.5 during pre-training.
For fine-tuning, we fine-tune our HOIL separately on Waymo~\cite{sun2020scalability} and SLOPER4D~\cite{dai2023sloper4d}.

\subsection{Evaluation metrics}
To evaluate 3D human pose estimation, we compute mean per joint position error~(MPJPE), percentage of correct keypoints with distance to GT less than 30\% of torso length~(PCK-3), percentage of correct keypoints with distance to GT less than
50\% of torso length~(PCK-5).

\vspace{-5mm}
\begin{table}[hbtp]
\centering
\footnotesize
\setlength{\tabcolsep}{2.7pt}
\caption{\textbf{Ablation of human-object interaction-aware contrastive learning on Waymo~\cite{sun2020scalability}.} 
CL refers to contrastive learning. 
}
\vspace{-3mm}
\begin{tabular}{@{}ccccccc@{}}
\toprule
Global CL & FIR CL & Contact CL & MPJPE~$\downarrow$ & PCK-3~$\uparrow$ & PCK-5~$\uparrow$ \\
\midrule
\ding{55} & \ding{55} & \ding{55} & 51.16 & 97.09 & 98.83 \\
\ding{51} & \ding{55} & \ding{55} & 50.05 & 97.21 & 98.87 \\
\ding{51} & \ding{51} & \ding{55} & 49.91 & 98.04 & 98.96 \\
\ding{51} & \ding{51} & \ding{51} &  \textbf{48.83} & \textbf{98.51} & \textbf{99.14} \\
\bottomrule
\end{tabular}
\label{tab:abl_contrastive}
\end{table}
\vspace{-13mm}

\begin{table}[htbp]
\centering
\footnotesize
\setlength{\tabcolsep}{4pt}
\caption{\textbf{Quantitative comparison of various contrastive learning techniques on Waymo~\cite{sun2020scalability}.} Baseline is a cross-entropy~(CE) loss without contrastive learning.}
\vspace{-3mm}
\begin{tabular}{@{}lccc@{}}
\toprule
Methods & MPJPE~$\downarrow$ & PCK-3~$\uparrow$ & PCK-5~$\uparrow$ \\ 
\midrule
Baseline~(CE)  & 51.16 & 97.09 & 98.83 \\
SupCon~\cite{khosla2020supervised} & 50.07 & 97.29 & 98.83 \\
KCL~\cite{kang2020exploring} & 50.03 & 97.31 & 98.89 \\
BCL~\cite{zhu2022balanced} & 49.92 & 97.51 & 98.81 \\
CBL~\cite{tang2022contrastive} & 51.13 & 97.10 & 98.73 \\
HiMulConE~\cite{zhang2022use} & 49.90 & 97.35 & 98.85 \\
TSC~\cite{li2022targeted} & 49.73 & 97.72 & 98.91 \\
HOICL~(Ours) & \textbf{48.83} & \textbf{98.51} & \textbf{99.14} \\
\bottomrule
\end{tabular}
\label{tab:sota_contrastive}
\end{table}
\vspace{-12mm}

\subsection{Ablation study}
\noindent\textbf{Effectiveness of human-object interaction-aware contrastive learning.}
Table~\ref{tab:abl_contrastive} demonstrates that human–object interaction-aware contrastive learning~(HOICL) improves performance on Waymo~\cite{sun2020scalability} dataset.
All components of HOICL, namely Global CL, FIR CL, Contact CL, provide consistent improvements.
In particular, the Contact CL shows considerable gains, indicating that contact regions suffer significantly from spatial ambiguity.
Additionally, Table~\ref{tab:sota_contrastive} presents a comparison of various supervised contrastive learning~(SCL) techniques~\cite{khosla2020supervised, kang2020exploring, zhu2022balanced, tang2022contrastive, zhang2022use, li2022targeted}.
The results show that HOICL achieves superior performance compared to other SCL methods.
This demonstrates that HOICL effectively resolves spatial ambiguity between human and object points, thereby improving 3D human pose estimation from LiDAR point clouds.

\noindent\textbf{Effectiveness of contact-aware part-guided pooling.}
Table~\ref{tab:abl_part_guided} presents an ablation study of contact-aware part-guided pooling~(CPPool).
As shown in the table, performance improves consistently as part and contact are incorporated into CPPool.
In particular, the largest gain in MPJPE is observed when contact is added, with a 2.32\% improvement.
Such improvement from contact is attributed to the severe class imbalance issue in contacting body parts.
With both part and contact, the PCK-5 exceeds 99\%.
These results demonstrate that CPPool effectively addresses the class imbalance issue.

\noindent\textbf{Effectiveness of contact-based temporal refinement.}
Table~\ref{tab:temporal} presents an ablation study of the contact-based temporal refinement module (CTRefine).
The results show that both temporal and contact cues improve the 3D human pose estimation performance of HOIL.
Note that the setting without temporal and contact corresponds to the original HOIL, which processes LiDAR points on a per-frame basis.
These results indicate that refining 3D human keypoint estimates using contact cues is effective, as contacting body parts are particularly challenging to estimate keypoints.
Our optional CTRefine effectively leverages temporal contact cues to refine erroneous keypoint predictions.

\vspace{-5mm}
\begin{table}[hbtp]
\centering
\footnotesize
\setlength{\tabcolsep}{2.7pt}
\caption{\textbf{Ablation of contact-aware part-guided pooling on Waymo~\cite{sun2020scalability}.}}
\vspace{-3mm}
\begin{tabular}{@{}cccccc@{}}
\toprule
Part-guided & Contact-aware & MPJPE~$\downarrow$ & PCK-3~$\uparrow$ & PCK-5~$\uparrow$ \\
\midrule
\ding{55} & \ding{55} & 50.78 & 97.03 & 98.79 \\
\ding{51} & \ding{55} & 49.99 & 97.38 & 98.96  \\
\ding{51} & \ding{51} & \textbf{48.83} & \textbf{98.51} & \textbf{99.14}  \\
\bottomrule
\end{tabular}
\label{tab:abl_part_guided}
\end{table}
\vspace{-10mm}

\begin{table}[htbp]
\centering
\footnotesize
\setlength{\tabcolsep}{4pt}
\caption{\textbf{Ablation of contact-based temporal refinement on InterCap~\cite{huang2024intercap}.} Note that temporal refinement is only conducted in this experiment.}
\vspace{-3mm}
\begin{tabular}{@{}ccccc@{}}
\toprule
Temporal & Contact & MPJPE~$\downarrow$ & PCK-3~$\uparrow$ & PCK-5~$\uparrow$ \\ 
\midrule
\ding{55} & \ding{55} & 38.87 & 97.52 & 99.33 \\
\ding{51} & \ding{55} & 37.82 & 97.61 & 99.40 \\
\ding{51} & \ding{51} & \textbf{36.69} & \textbf{97.91} & \textbf{99.60} \\
\bottomrule
\end{tabular}
\label{tab:temporal}
\end{table}
\vspace{-7mm}

\subsection{Comparison with state-of-the-art methods}

\noindent\textbf{Quantitative results.}
Table~\ref{tab:sota_comparison_hpe} presents a comparison between our HOIL and SOTA methods, including NE~\cite{zhang2024neighborhood}, LPFormer~\cite{ye2024lpformer}, PRN~\cite{fan2025lidar}, and DAPT~\cite{an2025pre}, on Waymo~\cite{sun2020scalability} and SLOPER4D~\cite{dai2023sloper4d} datasets.
Our method achieves meaningful improvements across all metrics on Waymo dataset, which is the most realistic LiDAR dataset with multiple real-world challenges such as diverse human–object interactions.
All methods exhibit inferior performance with large errors in either human–object interaction scenarios~(\textit{e.g.,} cycling, walking with an umbrella) or novel body pose scenarios~(\textit{e.g.,} sitting on the ground).
By effectively handling human–object interaction scenarios, our method achieves superior performance over prior state-of-the-art methods, demonstrating its potential as a strong model for 3D human pose estimation from LiDAR point clouds.

\begin{figure}[t]
\begin{center}
\includegraphics[width=1.0\linewidth]{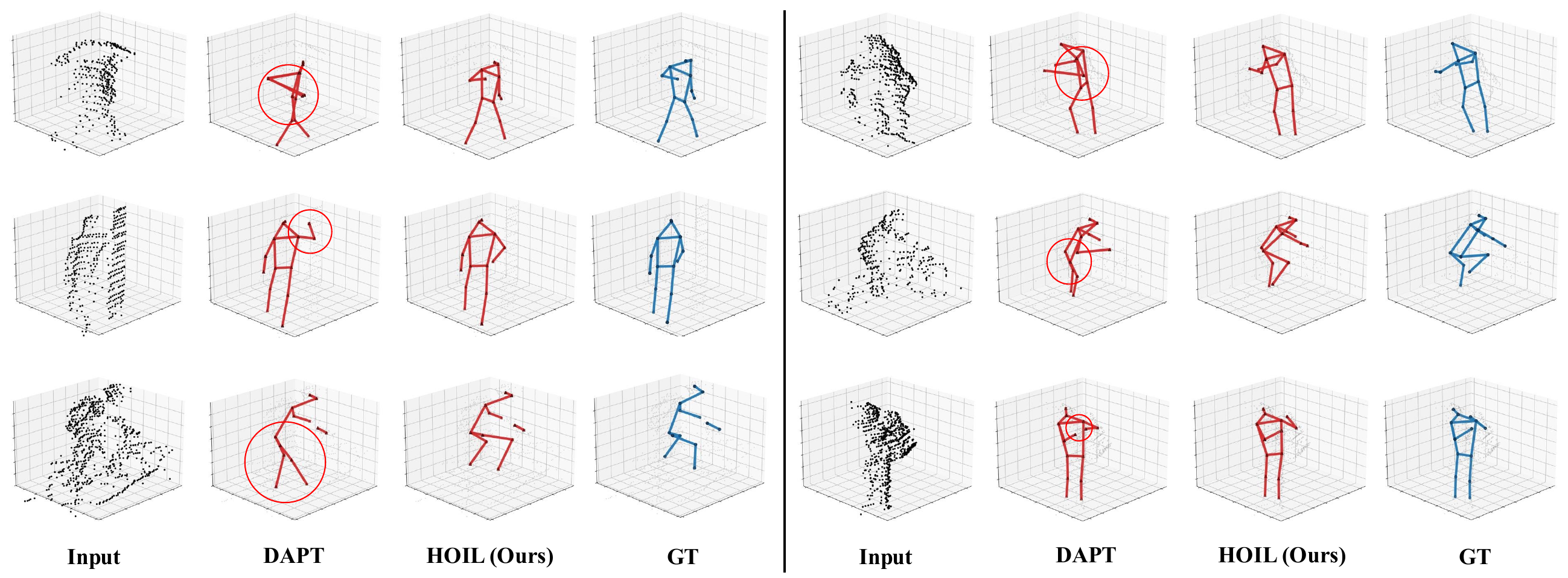}
\end{center}
\vspace{-4.5mm}
\caption{
\textbf{Qualitative comparison of 3D human pose estimation with the state-of-the-art method~\cite{an2025pre} on Waymo~\cite{sun2020scalability}.} Red circles indicate exemplar regions that HOIL outperforms previous methods.}
\label{fig:qual_3d_hpe}
\vspace{-0.3cm}
\end{figure}

\begin{table}[htbp]
\centering
\small
\setlength{\tabcolsep}{4pt}
\caption{\textbf{Quantitative comparison with SOTA methods of 3D human pose estimation on Waymo~\cite{sun2020scalability} and SLOPER4D~\cite{dai2023sloper4d}.}}
\vspace{-3mm}
\begin{tabular}{llccc}
\toprule
Test dataset & Methods & MPJPE~$\downarrow$ & PCK-3~$\uparrow$ & PCK-5~$\uparrow$ \\
\midrule
\multirow{4}{*}{Waymo~\cite{sun2020scalability}} 
& NE~\cite{zhang2024neighborhood} & -- & -- & -- \\
& LPFormer~\cite{ye2024lpformer} & 61.60 & 94.52 & 98.04 \\
& PRN~\cite{fan2025lidar} & 68.48 & 93.60 & 97.87 \\
& DAPT~\cite{an2025pre} & 51.59 & 97.34 & 98.98 \\
& HOIL~(Ours) & \textbf{48.83} & \textbf{98.51} & \textbf{99.14} \\
\midrule
\multirow{4}{*}{SLOPER4D~\cite{dai2023sloper4d}} 
& NE~\cite{zhang2024neighborhood} & 96.80 &  79.22 &  90.51 \\
& LPFormer~\cite{ye2024lpformer} & 49.31 & 97.22 & 99.44 \\
& PRN~\cite{fan2025lidar} & 48.76 & 97.38 & 99.45 \\
& DAPT~\cite{an2025pre} & 28.01 & 99.30 & 99.87 \\
& HOIL~(Ours) & \textbf{22.68} & \textbf{99.67} & \textbf{99.91} \\
\bottomrule
\end{tabular}
\label{tab:sota_comparison_hpe}
\vspace{-0.4cm}
\end{table}

\noindent\textbf{Qualitative results.}
Fig.~\ref{fig:qual_3d_hpe} shows that HOIL effectively learns human–object interactions compared to DAPT~\cite{an2025pre}, the SOTA method, on the Waymo~\cite{sun2020scalability} dataset.
In particular, we evaluate human–object interaction samples involving an umbrella~(first row), a wall~(second row), and a bicycle~(third and fifth rows).
In the first row, DAPT fails due to the umbrella, which causes spatial ambiguity between human and object points, whereas HOIL produces accurate pose.
In the second row, DAPT incorrectly predicts the human raising the right arm, whereas HOIL correctly places the right hand in a natural position near the hip.
In the third and fifth rows, the results clearly show that DAPT incorrectly predicts the cyclist’s pose as standing, whereas HOIL successfully predicts a sitting pose.
Overall, our HOIL demonstrates robust and accurate human–object interaction understanding in the presence of surrounding objects by addressing key issues.
\section{Conclusion}
\label{sec:conclusion}
We propose HOIL, a human-object interaction learning framework for 3D human pose estimation from LiDAR point clouds.
To address spatial ambiguity and class imbalance in human-object interaction regions, we introduce human-object interaction-aware contrastive learning and contact-aware part-guided pooling.
As a result, HOIL achieves robust and accurate pose estimation in human-object interaction scenarios, outperforming prior methods by a significant margin.

\clearpage

\begin{center}
\textbf{\large Supplementary Material \textit{for} \\ \vspace{2mm}
\large{``Learning Human-Object Interaction\\for 3D Human Pose Estimation\\from LiDAR Point Clouds"}}
\end{center}

\setcounter{section}{0}
\setcounter{table}{0}
\setcounter{figure}{0}

\renewcommand{\thesection}{S\arabic{section}}
\renewcommand{\thetable}{S\arabic{table}}   
\renewcommand{\thefigure}{S\arabic{figure}}

\renewcommand{\theHsection}{S\arabic{section}}
\renewcommand{\theHtable}{S\arabic{table}}
\renewcommand{\theHfigure}{S\arabic{figure}}

In this supplementary material, we present additional experiments, discussions and other results that could not be included in the main manuscript due to the lack of pages.
The contents are summarized below:
\begin{compactitem}
    \item \ref{sec:supp_loss_function}. Details of loss functions
    \item \ref{sec:supp_part_def}. Details of keypoint and part definition
    \item \ref{sec:supp_keypoint_embeddings}. Details of keypoint query embeddings
    \item \ref{sec:supp_contact_labels}. Details of contact labels
    \item \ref{sec:supp_spatial_amb}. Analysis on spatial ambiguity issue
    \item \ref{sec:supp_class_imb}. Analysis on class imbalance issue
    \item \ref{sec:supp_quant_ctrefine}. Quantitative results of CTRefine
    \item \ref{sec:supp_comp_require}. Computational requirements
    \item \ref{sec:supp_more_qual}. More qualitative results
    \item \ref{sec:supp_limit_and_societal}. Limitations and societal impacts
\end{compactitem}

\section{Details of loss functions}
\label{sec:supp_loss_function}

\noindent\textbf{CPPool loss.}
As described in the main manuscript, CPPool introduces three predictors inside the pooling module: an auxiliary part segmentation head~$f_{\mathrm{part}}$, an auxiliary contact head~$f_{\mathrm{contact}}$, and an auxiliary importance head~$f_{\mathrm{imp}}$. Our CPPool loss is defined as:
\begin{equation}
\mathcal{L}_{\mathrm{CPPool}}
=
\mathcal{L}_{\mathrm{cppool\_part}}
+
\mathcal{L}_{\mathrm{cppool\_contact}} .
\end{equation}
The loss $\mathcal{L}_{\mathrm{cppool\_part}}$ supervises the part probabilities predicted by $f_{\mathrm{part}}$ using GT part segmentation labels with a CE loss. Similarly, $\mathcal{L}_{\mathrm{cppool\_contact}}$ supervises the point-level contact prediction from $f_{\mathrm{contact}}$ using GT contact labels with a CE loss. 
The importance head $f_{\mathrm{imp}}$ predicts importance logits that contribute to the pooling logits used for weighted feature aggregation in CPPool. 
The parameters of $f_{\mathrm{imp}}$ are learned implicitly through backpropagation from the overall training objective of HOIL. 
In our implementation, we set $\lambda_{\mathrm{part}}=1.0$ and $\lambda_{\mathrm{contact}}=1.0$ when computing the final pooling logit~$\boldsymbol{\ell}$.

\noindent\textbf{Limb loss.}
Motivated by PoseConv3D~\cite{duan2022revisiting}, we introduce a limb regularization term that encourages kinematically consistent skeleton structures during fine-tuning. 
Let $a$ and $b$ denote two arbitrary joints connected by a bone $(a,b)$ in the skeleton, and let $\mathbf{K}_a$ and $\mathbf{K}_b$ denote the predicted 3D keypoint coordinates of joints $a$ and $b$, respectively. 
The predicted and ground-truth bone vectors are defined as:
\begin{equation}
\mathbf{B}^{\mathrm{pred}}_{ab} = \mathbf{K}^{\mathrm{pred}}_b - \mathbf{K}^{\mathrm{pred}}_a,
\qquad
\mathbf{B}^{\mathrm{gt}}_{ab} = \mathbf{K}^{\mathrm{gt}}_b - \mathbf{K}^{\mathrm{gt}}_a.
\end{equation}
The limb loss penalizes both bone direction and bone length differences, where the direction term~$\mathcal{L}_{\mathrm{dir}}$ measures cosine similarity between predicted and ground-truth bone vectors and the length term~$\mathcal{L}_{\mathrm{len}}$ applies a SmoothL1 loss to the length magnitudes. 
The final limb loss is defined as:
\begin{equation}
\mathcal{L}_{\mathrm{limb}}
=
\lambda_{\mathrm{dir}} \mathcal{L}_{\mathrm{dir}}
+
\lambda_{\mathrm{len}} \mathcal{L}_{\mathrm{len}},
\end{equation}
where we set $\lambda_{\mathrm{dir}}=1.0$ and $\lambda_{\mathrm{len}}=1.0$. 

\noindent\textbf{Loss weights.}
For pre-training, the total loss consists of the segmentation loss $\mathcal{L}_{\mathrm{seg}}$, contact loss $\mathcal{L}_{\mathrm{contact}}$, 3D keypoint coordinate loss $\mathcal{L}_{\mathrm{coord}}$, keypoint-level contact loss $\mathcal{L}_{\mathrm{K\_contact}}$, HOI contrastive loss $\mathcal{L}_{\mathrm{HOICL}}$, and CPPool loss $\mathcal{L}_{\mathrm{CPPool}}$. The corresponding loss weights are set to $1.0$, $1.0$, $0.5$, $0.02$, $1.0$, and $1.0$, respectively. 
For the segmentation head used in pre-training, we combine CE loss with contrastive objectives, where $\lambda_{\mathrm{hmlc}}=0.05$ and $\lambda_{\mathrm{tsc}}=0.05$ denote the weights for $\mathcal{L}_{\mathrm{HMLC}}$ and $\mathcal{L}_{\mathrm{TSC}}$, respectively. 
For the HOICL, we set $\lambda_{\mathrm{fir}}=1.0$ and $\lambda_{\mathrm{hoc}}=1.0$ as the weights for the feature interaction regularization and HOI contrastive components, respectively. 
During fine-tuning, the optimization objective consists of the heatmap loss $\mathcal{L}_{\mathrm{heatmap}}$ and the limb loss $\mathcal{L}_{\mathrm{limb}}$, whose weights are set to $1.0$ and $0.1$, respectively.

\section{Details of keypoint and part definition}
\label{sec:supp_part_def}

\noindent\textbf{Keypoint definition.}
Following DAPT~\cite{an2025pre}, our framework mostly uses SMPL human body joints to maintain consistency across datasets.
For datasets providing human body meshes~(\textit{e.g.,} InterCap, SLOPER4D), 3D keypoints are obtained from SMPL joints regressed from the SMPL mesh vertices, whereas for datasets that provide direct keypoint annotations~(\textit{e.g.,} Waymo), we follow the dataset-provided keypoint definition.
Since different datasets contain different joint conventions and annotation coverage, we use dataset-specific subsets of keypoints during fine-tuning.
During pre-training, we use five human-object interaction datasets~\cite{bhatnagar2022behave, jiang2023full, zhang2023neuraldome, li2023object, huang2024intercap} that provide SMPL meshes and therefore supervise the model using SMPL joints.
The SMPL skeleton contains 24 human joints, from which we follow DAPT~\cite{an2025pre} and use a subset of 15 keypoints covering major body joints such as hips, knees, ankles, shoulders, elbows, wrists, neck, and head.
For human-object interaction datasets during pre-training, we additionally include one object keypoint defined as the centroid of the object mesh, resulting in a total of 16 keypoints.
For Waymo, we use the subset of annotated body joints provided by the dataset, and after removing non-semantic entries the resulting supervision consists of 14 keypoints corresponding to the head, shoulders, elbows, wrists, hips, knees, and ankles.

\noindent\textbf{Kinematic structure.}
For each dataset, the kinematic tree is defined in the original annotation space and then remapped to the subset of keypoints .
Edges associated with excluded joints, namely joints that are not annotated in the ground truth, are removed so that the skeleton structure remains valid for the reduced keypoint set.
This skeleton definition is used by the limb loss~$\mathcal{L}_{\mathrm{limb}}$.

\noindent\textbf{Part definition.}
During pre-training with human-object interaction datasets, we supervise point-level segmentation.
The SMPL mesh faces are assigned semantic body-part labels using a predefined SMPL part mapping.
Each face is associated with one human body-part label, while all object faces are assigned to a single object label.
Points that do not belong to either the human mesh or the object mesh, such as ground or wall points, are assigned a background label.
Consequently, the segmentation task consists of 24 human body-part classes, one object class, and one background class.

\section{Details of keypoint query embeddings}
\label{sec:supp_keypoint_embeddings}

Following DAPT~\cite{an2025pre} and the keypoint definitions described in Section~\ref{sec:supp_part_def}, we use learnable keypoint queries to represent the target keypoints.
Each query is initialized from a learnable embedding in which one token corresponds to one keypoint.
The number of query embeddings therefore matches the number of keypoints in the corresponding training setup.
During pre-training with human-object interaction datasets, the model uses 16 queries corresponding to 15 SMPL body joints and one object keypoint, while during fine-tuning the query embeddings are re-initialized to match the dataset-specific keypoint definitions (\textit{e.g.,} 14 keypoints for Waymo).
The query embedding dimension is 256 in all experiments.
For each input sample in a mini-batch, the same learnable embeddings are repeated across the batch dimension to construct the initial keypoint queries.
Although the number of keypoint queries differ between pre-training and fine-tuning due to the different keypoint sets, the overall mechanism remains the same across pre-training and fine-tuning, where the queries attend to point features through the encoder-decoder architecture.

\section{Details of contact labels}
\label{sec:supp_contact_labels}

Here, we explain how contact labels are obtained from the human-object interaction datasets~\cite{bhatnagar2022behave, jiang2023full, zhang2023neuraldome, li2023object, huang2024intercap} used for point-level and keypoint-level contact supervision.
These datasets provide paired human and object meshes, which allow us to extract dense human-object contact labels.

\noindent\textbf{Point-level contact.}
For the human-object interaction datasets used in pre-training, we first extract dense human-object contact labels on the mesh using distance-based thresholding, following prior works~\cite{nam2024joint, jung2025learning, jung2025shoe} on dense contact estimation.
Specifically, after aligning the human and object meshes to a common coordinate system, we compute the distance from each human mesh vertex to the object surface and from each object mesh vertex to the human surface using proximity queries.
A mesh vertex is labeled as contact if its distance to the other surface is smaller than a fixed threshold of \(5\)~cm.
The resulting binary vertex-level contact labels are then converted to face-level contact labels.
Point-level contact labels are not extracted directly from the mesh.
Instead, after simulating LiDAR points by ray casting, each point is assigned the contact label of the mesh face from which it is sampled using the point-to-face correspondence.
In the end, point-level contact is defined only for points sampled from the human or object mesh.
Background points on ground or wall are not considered and are not assigned human-object contact labels.

\noindent\textbf{Keypoint-level contact.}
Keypoint-level contact labels are derived from the vertex-level contact annotations.
After regressing SMPL joints from the aligned SMPL mesh, we determine the contact state of each keypoint by aggregating the contact states of the mesh vertices associated with the corresponding body part.
A keypoint is labeled as contact if at least one of its associated vertices is labeled as contact.
For the object, the keypoint-level contact label is determined by aggregating the contact states of the object mesh vertices.
This produces binary keypoint-level contact labels for the human keypoints and the object keypoint.
For real-world LiDAR datasets used during fine-tuning, explicit mesh-based contact annotations are not available.
Instead, we obtain keypoint-level contact labels using a zero-velocity constraint heuristic used in WHAM~\cite{shin2024wham}.
Specifically, joints whose velocities remain below a predefined threshold across consecutive frames are treated as contact joints.
These estimated labels are used as supervision for keypoint-level contact prediction during training, which is required for CTRefine.
In Table~\ref{tab:supp_ctrefine_sloper}, we show experiments of HOIL on SLOPER4D that is trained with keypoint-level contact extracted with the zero-velocity constraint heuristic.

\begin{figure}[htbp]
\begin{center}
\includegraphics[width=0.7\linewidth]{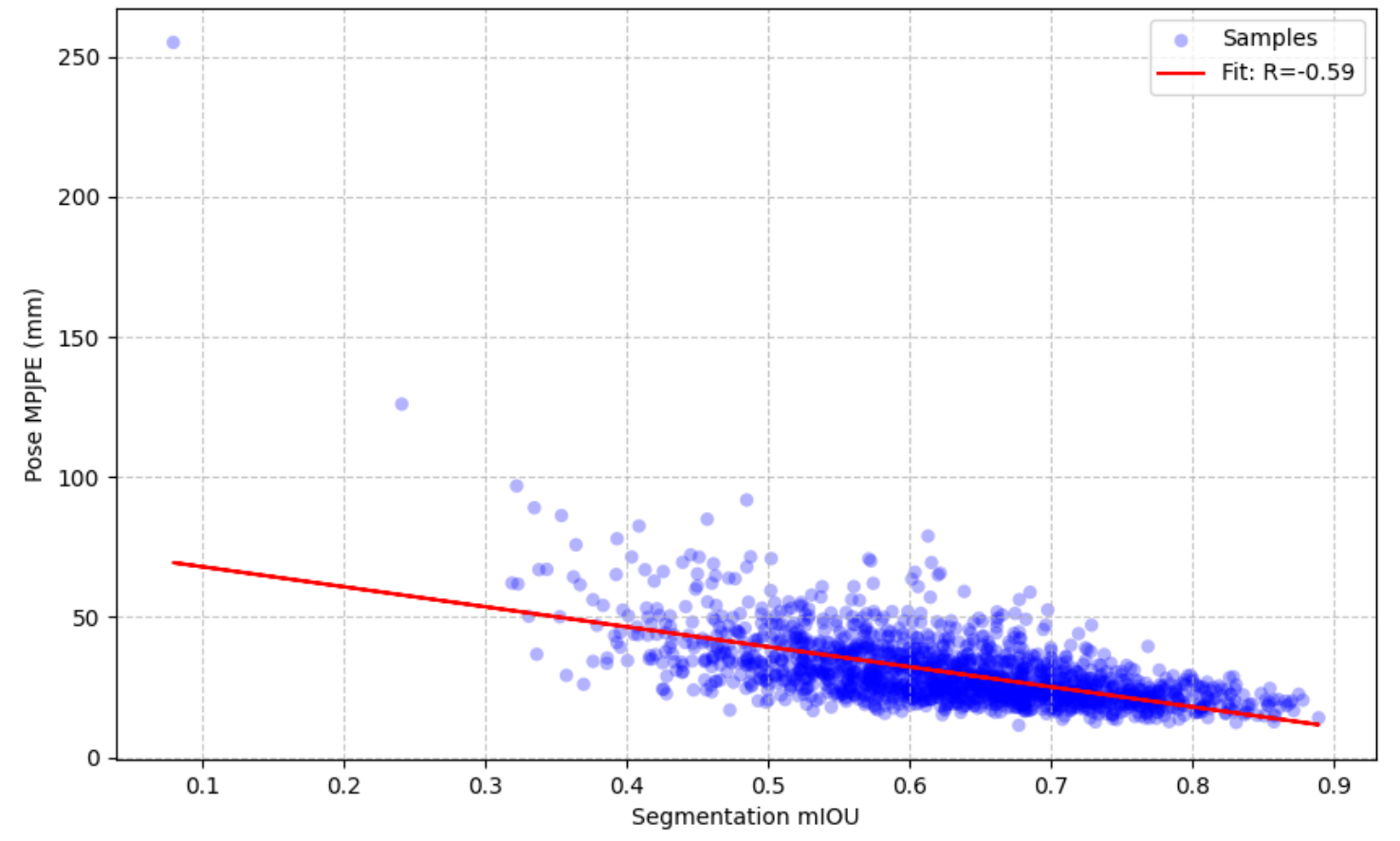}
\end{center}
\vspace{-6mm}
\caption{
\textbf{Correlation between segmentation accuracy and human pose estimation error of state-of-the-art method~\cite{an2025pre} on Waymo~\cite{sun2020scalability}.}}
\label{fig:supp_corr_seg_pose}
\end{figure}
\vspace{-6mm}

\section{Analysis on spatial ambiguity issue}
\label{sec:supp_spatial_amb}
In the main manuscript, we posed the spatial ambiguity issue between human and object points as one of the two main challenges for 3D human pose estimation from LiDAR points.
To support this claim, we analyze the spatial ambiguity issue by examining the correlation between segmentation accuracy and pose error of the previous state-of-the-art method~\cite{an2025pre}.
Figure~\ref{fig:supp_corr_seg_pose} shows that there is a meaningful negative correlation between segmentation accuracy and human pose estimation error.
Specifically, we perform linear regression and observe a correlation coefficient of $R=-0.59$.
This indicates that the ability to accurately segment and differentiate points belonging to different human body parts is important for accurate 3D human pose estimation from LiDAR point clouds.

\begin{figure}[htbp]
\begin{center}
\includegraphics[width=1.0\linewidth]{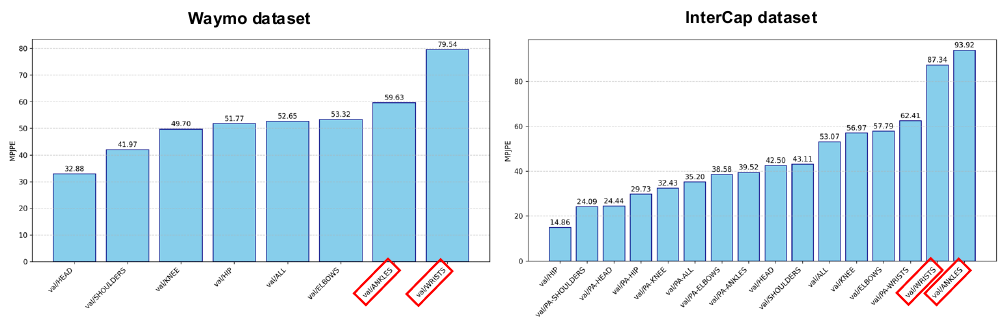}
\end{center}
\vspace{-6mm}
\caption{
\textbf{Error by human body joints of state-of-the-art method~\cite{an2025pre} on Waymo~\cite{sun2020scalability} and InterCap~\cite{huang2024intercap}.} Red rectangles indicate errors from wrists and ankles which correspond to frequently interacting parts~(\textit{e.g.,} hands and feet).}
\label{fig:supp_error_body_part}
\end{figure}
\vspace{-6mm}

\section{Analysis on class imbalance issue}
\label{sec:supp_class_imb}
In the main paper, we reported the number of points for human, object, and background parts and showed that there exists a class imbalance issue for frequently interacting body parts compared to other body parts.
Here, we show that the frequently interacting body parts that suffer from the class imbalance issue are also the human body joints that suffer from the highest errors in the state-of-the-art method~\cite{an2025pre}.
We can clearly observe that the frequently interacting body part joints correspond to the human body joints with the highest errors.
For instance, in the Waymo dataset~\cite{sun2020scalability}, the wrist error is significantly higher than that of the human body joint with the highest error among non-frequently interacting joints (\textit{i.e.}, elbow) by 32.96\%.
Moreover, in the InterCap dataset~\cite{huang2024intercap}, the ankle error is 38.46\% higher than the elbow error, which is also the non-frequently interacting human body joint with the highest error.

\section{Quantitative results of CTRefine}
\label{sec:supp_quant_ctrefine}

Table~\ref{tab:supp_comp_ctrefine} presents a quantitative comparison of CTRefine with several commonly used temporal smoothing techniques on the InterCap~\cite{huang2024intercap} dataset, including Gaussian filtering~\cite{young1995recursive}, Savitzky--Golay filtering~\cite{press1990savitzky}, and the One-Euro filtering~\cite{casiez20121}, which are widely used for stabilizing temporal predictions. 
As shown in Table~\ref{tab:supp_comp_ctrefine}, CTRefine consistently outperforms these filtering-based approaches and achieves approximately a 5.6\% reduction in MPJPE compared to the baseline without refinement while also improving the PCK metrics, suggesting that CTRefine effectively leverages temporal information for pose refinement beyond simple signal smoothing techniques. 
Table~\ref{tab:supp_ctrefine_sloper} further presents an ablation study of CTRefine on the SLOPER4D~\cite{dai2023sloper4d} dataset, where incorporating CTRefine leads to consistent improvement in pose estimation accuracy. 
Although the improvement is modest due to the already strong baseline performance and the supervision of heuristic contact ground-truth during training, the results demonstrate that CTRefine provides stable temporal refinement without degrading pose estimation performance. 
Note that the subset of SLOPER4D used in Table~\ref{tab:supp_ctrefine_sloper} differs from that of the main paper, as the main paper evaluates frame-based samples while Table~\ref{tab:supp_ctrefine_sloper} evaluates video-based samples.

\begin{table}[htbp]
\centering
\footnotesize
\setlength{\tabcolsep}{4pt}
\caption{\textbf{Quantitative comparison of various temporal refinement techniques on InterCap~\cite{huang2024intercap}.}}
\vspace{-3mm}
\begin{tabular}{@{}cccc@{}}
\toprule
Methods & MPJPE~$\downarrow$ & PCK-3~$\uparrow$ & PCK-5~$\uparrow$ \\ 
\midrule
No refinement & 38.87 & 97.52 & 99.33 \\
Gaussian filter~\cite{young1995recursive} & 37.30 & 97.90 & 99.58 \\
Savitzky-Golay filter~\cite{press1990savitzky} & 37.08 & 97.84 & 99.53 \\
One-Euro filter~\cite{casiez20121} & 38.52 & 97.56 & 99.36 \\
CTRefine~(Ours) & \textbf{36.69} & \textbf{97.91} & \textbf{99.60} \\ 
\bottomrule
\end{tabular}
\label{tab:supp_comp_ctrefine}
\end{table}
\vspace{-6mm}

\begin{table}[htbp]
\centering
\footnotesize
\setlength{\tabcolsep}{4pt}
\caption{\textbf{Ablation of CTRefine on SLOPER4D~\cite{dai2023sloper4d}.}}
\vspace{-3mm}
\begin{tabular}{@{}cccc@{}}
\toprule
Methods & MPJPE~$\downarrow$ & PCK-3~$\uparrow$ & PCK-5~$\uparrow$ \\ 
\midrule
w/o CTRefine & 25.43 & 98.67 & \textbf{99.65} \\
w/ CTRefine~(Ours) & \textbf{25.36} & \textbf{98.69} & \textbf{99.65} \\
\bottomrule
\end{tabular}
\label{tab:supp_ctrefine_sloper}
\end{table}
\vspace{-6mm}

\section{Computational requirements}
\label{sec:supp_comp_require}

Table~\ref{tab:supp_comp_require} reports the computational requirements of HOIL, including memory consumption, model size, inference speed, and computational cost.
The proposed model contains 53.02M parameters.
During training, HOIL requires 33,198MB and 20,262MB GPU memory for the pre-training and fine-tuning stages, respectively.
At inference time, the model consumes 8,435MB GPU memory and runs at 155.52~Hz, corresponding to approximately 6.43~ms per point cloud.
The computational complexity of the model is 393.89 GFLOPs measured with a single forward pass.
These results demonstrate that HOIL maintains reasonable computational performance while supporting 3D human pose estimation.

\begin{table}[htbp]
\centering
\scriptsize
\setlength{\tabcolsep}{3pt}
\caption{\textbf{Computational requirements of HOIL.}}
\vspace{-3mm}
\begin{tabular}{lcccccc} \toprule
Model & Pre-train Memory & Fine-tune Memory & Test Memory & Params. & Speed & GFLOPs \\
\midrule
HOIL & 33,198MB & 20,262MB & 8,435MB & 53.02M & 155.52 Hz & 393.89 \\
\bottomrule
\end{tabular}
\vspace{-0.3cm}
\label{tab:supp_comp_require}
\end{table}

\section{More qualitative results}
\label{sec:supp_more_qual}

\begin{figure}[htbp]
\begin{center}
\includegraphics[width=0.9\linewidth]{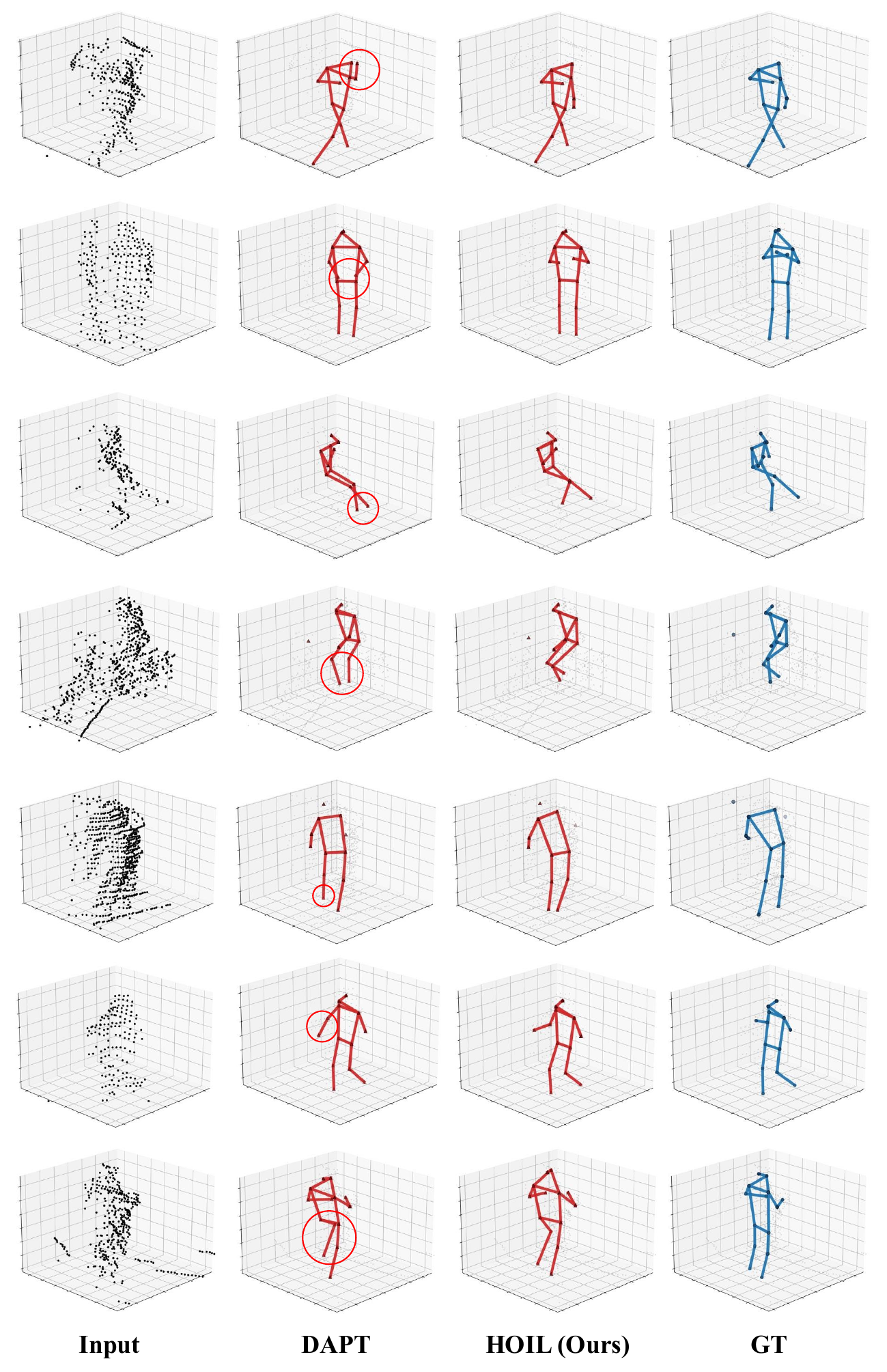}
\end{center}
\vspace{-4.5mm}
\caption{
\textbf{Qualitative comparison of 3D human pose estimation with the state-of-the-art method~\cite{an2025pre} on Waymo~\cite{sun2020scalability}.} We assort samples where the SOTA method has highest errors from the dataset. Red circles indicate exemplar regions that HOIL outperforms previous methods.}
\label{fig:supp_sota_hpe_qual}
\end{figure}

\begin{figure*}[htbp]
\begin{center}
\includegraphics[width=0.9\linewidth]{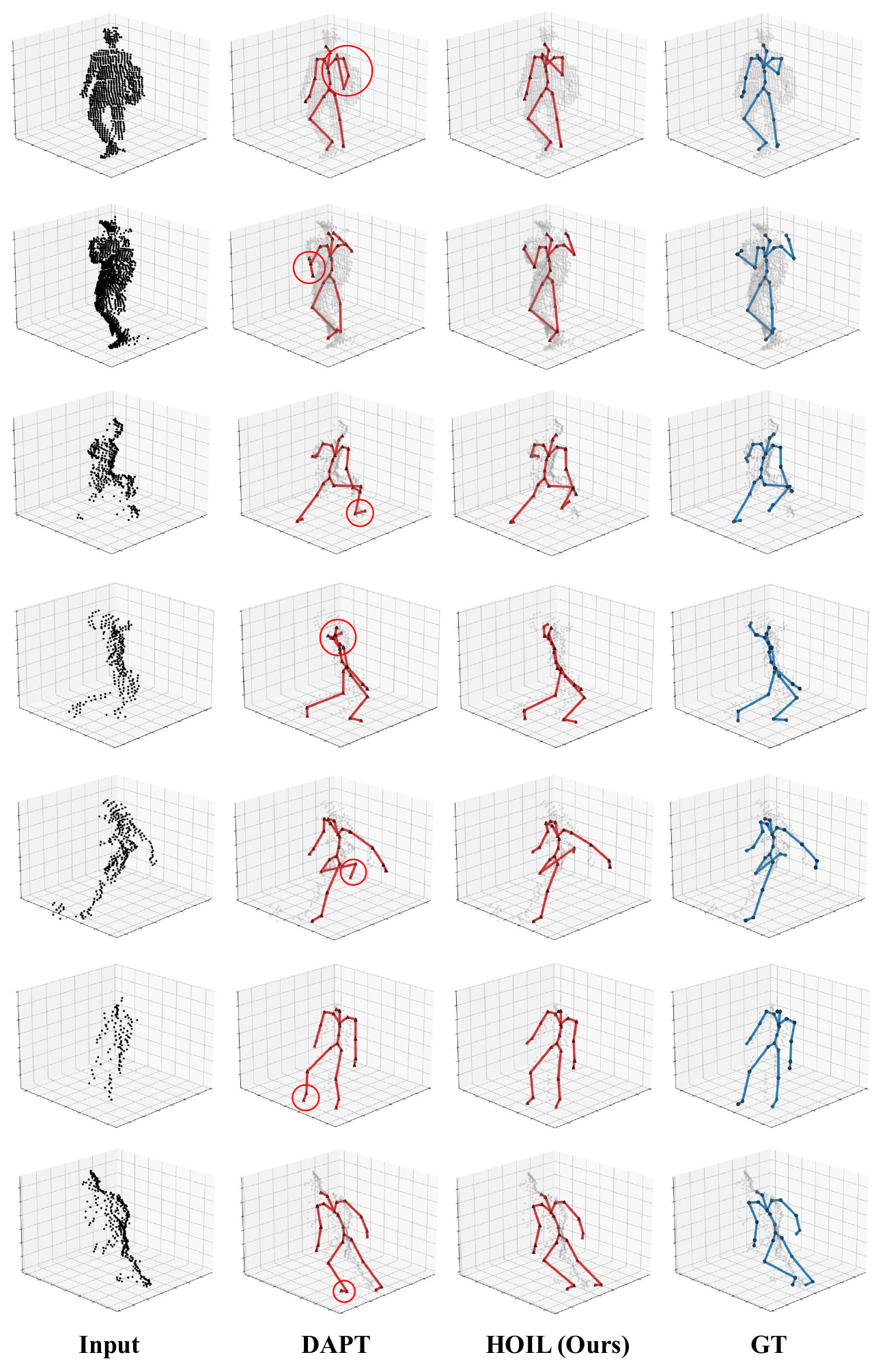}
\end{center}
\vspace{-4.5mm}
\caption{
\textbf{Qualitative comparison of 3D human pose estimation with the state-of-the-art method~\cite{an2025pre} on SLOPER4D~\cite{dai2023sloper4d}.} We assort samples where the SOTA method has highest errors from the dataset. Red circles indicate exemplar regions that HOIL outperforms previous methods.}
\label{fig:supp_sota_hpe_qual_sloper}
\end{figure*}

In Figure~\ref{fig:supp_sota_hpe_qual}, we present qualitative comparisons of 3D human pose estimation with the state-of-the-art method~\cite{an2025pre} on the Waymo~\cite{sun2020scalability} dataset.
We visualize the outputs of the previous SOTA method and our method on particularly challenging samples.
The first row illustrates a human-object interaction scenario where a person is walking with an umbrella.
Due to the spatial ambiguity introduced by the human-object interaction, DAPT predicts an incorrect right-hand keypoint location, whereas our HOIL correctly places the right-hand keypoint near the right hip.
The fourth row presents another challenging interaction scenario where a person is riding a motorcycle, in which our method produces a more realistic human pose that is closer to the ground-truth pose.
Figure~\ref{fig:supp_sota_hpe_qual_sloper} provides additional qualitative comparisons with the state-of-the-art method~\cite{an2025pre} on the SLOPER4D~\cite{dai2023sloper4d} dataset.
The first row depicts a person carrying a backpack, where HOIL predicts more accurate hand keypoints than DAPT.
The third row corresponds to a scenario where a person is playing soccer with a ball near the right foot, and our method produces a more accurate prediction of the right foot keypoint.
Overall, our method demonstrates superior performance in human-object interaction scenarios, where two major challenges arise: the spatial ambiguity between human and object points and the class imbalance of interaction-related body parts.

\section{Limitations and societal impacts}
\label{sec:supp_limit_and_societal}

\noindent\textbf{Limitations.}
Our HOIL effectively learns human--object interactions from dedicated datasets to address two major challenges in LiDAR-based 3D human pose estimation: spatial ambiguity and class imbalance.
However, our current framework does not leverage RGB information paired with LiDAR, unlike several previous works~\cite{furst2021hperl, zheng2022multi, zanfir2023hum3dil, cong2023weakly}.
While LiDAR provides direct geometric cues of humans and surrounding scenes, it lacks the rich semantic and contextual information available in RGB images.
Furthermore, due to the radial scanning mechanism of LiDAR sensors, point density decreases with distance.
As a result, when a person is far from the sensor, only a few LiDAR points may be observed, sometimes approaching only a few points per body part.
In such scenarios, powerful visual encoders could provide complementary cues to approximate the local human pose, while sparse LiDAR points primarily contribute to estimating the global location of the person.
Another limitation lies in the diversity of human--object interactions available in the training datasets.
Although we leverage five independent human--object interaction datasets, certain important interactions remain underrepresented.
For instance, cycling and motorcycling are common human behaviors in roadside environments, yet these interactions are absent from the datasets used in our training.
Consequently, the model can only indirectly learn similar poses, such as seated postures, from available data.
Recent works~\cite{corral20253darticcyclists, corral2025monocular} have explored generating controllable human cycling data.
If such data can be produced with paired SMPL annotations, it could further be used to synthesize LiDAR observations for large-scale pre-training, potentially improving HOIL's performance on cycling-related human poses.

\noindent\textbf{Societal impacts.}
The proposed method may benefit a wide range of applications that require the understanding of human motion and interactions, including pedestrian intention prediction, collision avoidance systems, crowd analysis, human--robot collaboration, and VR/AR systems.
However, technologies that analyze human behavior also carry the potential risk of misuse, particularly in contexts such as large-scale monitoring or surveillance.
This work aims to support safety-critical and assistive technologies. 
Nevertheless, its deployment should carefully consider issues related to personal privacy and individual rights.
We encourage future work and deployments to carefully consider ethical implications when applying this technology in real-world applications.

\clearpage

%
%
\bibliographystyle{splncs04}
\bibliography{main}
\end{document}